\documentclass[runningheads]{llncs}
\usepackage[T1]{fontenc}
\usepackage{graphicx}
\usepackage{graphicx}
\usepackage{subfigure}
\usepackage{makecell}
\usepackage{multirow}
\usepackage{booktabs}
\usepackage{amssymb}
\usepackage{bm}
\usepackage{soul}
\usepackage{url}
\usepackage{threeparttable}
\usepackage{amsmath}
\usepackage{subcaption}
\usepackage{float}
\usepackage{hyperref}
\usepackage[misc]{ifsym}
\begin{document}
\title{SCFormer: Structured Channel-wise Transformer with Cumulative Historical State for Multivariate Time Series Forecasting}
\titlerunning{Structured Channel-wise Transformer with Cumulative Historical State}
%
\author{Shiwei Guo\inst{1,2,3} \and
Ziang Chen\inst{1,2,3} \and
Yupeng Ma\inst{1,3}\textsuperscript{(\Letter)} \and
Yunfei Han\inst{1,3} \and
Yi Wang\inst{1,3}}
\authorrunning{Shiwei Guo et al.}
\institute{\textsuperscript{1}Xinjiang Technical Institute of Physics and Chemistry, Chinese Academy of Sciences, Urumqi, China \\
\email{\{ypma,hanyf,wangyi\}@ms.xjb.ac.cn}\\
\textsuperscript{2}University of Chinese Academy of Sciences, Beijing, China\\
\email{\{guoshiwei18,chenziang21\}@mails.ucas.ac.cn}\\
\textsuperscript{3}Xinjiang Laboratory of Minority Speech and Language Information Processing, Urumqi, China}
\maketitle              
\begin{abstract}
The Transformer model has shown strong performance in multivariate time series forecasting by leveraging channel-wise self-attention. However, this approach lacks temporal constraints when computing temporal features and does not utilize cumulative historical series effectively. To address these limitations, we propose the \textbf{S}tructured \textbf{C}hannel-wise Transformer with Cumulative Historical state (SCFormer). SCFormer introduces temporal constraints to all linear transformations, including the query, key, and value matrices, as well as the fully connected layers within the Transformer. Additionally, SCFormer employs High-order Polynomial Projection Operators (HiPPO) to deal with cumulative historical time series, allowing the model to incorporate information beyond the look-back window during prediction. Extensive experiments on multiple real-world datasets demonstrate that SCFormer significantly outperforms mainstream baselines, highlighting its effectiveness in enhancing time series forecasting. The code is publicly available at \url{https://github.com/ShiweiGuo1995/SCFormer}

\end{abstract}
\section{Introduction}
The Transformer, a versatile sequence model, has been widely applied in various fields, including NLP~\cite{tetko2020state}, computer vision~\cite{han2021transformer}, and bioinformatics~\cite{zhang2023applications}. Transformer-based models have also achieved significant progress in time series forecasting~\cite{liu2022memory,zhou2022fedformer,zhang2022crossformer}. Notably, recent studies have demonstrated that channel-wise Transformers~\cite{liu2023itransformer,ilbert2024unlocking} can effectively capture relationships among multiple temporal variables, resulting in substantial reductions in prediction errors.

\begin{figure}[htbp]
    \centering
        \centering
        \subfigure[\scalebox{0.9}{}]{\includegraphics[scale=0.4]{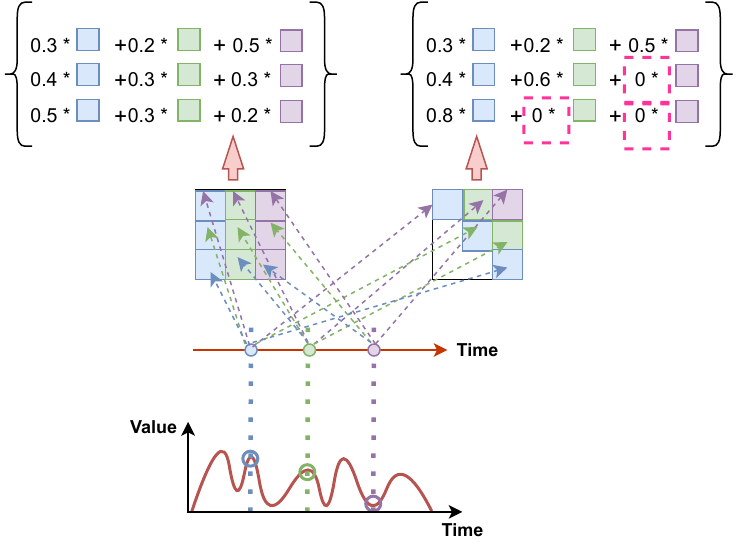}}
    \hfill 
        \centering
        \subfigure[\scalebox{0.9}{}]{\includegraphics[scale=0.28]{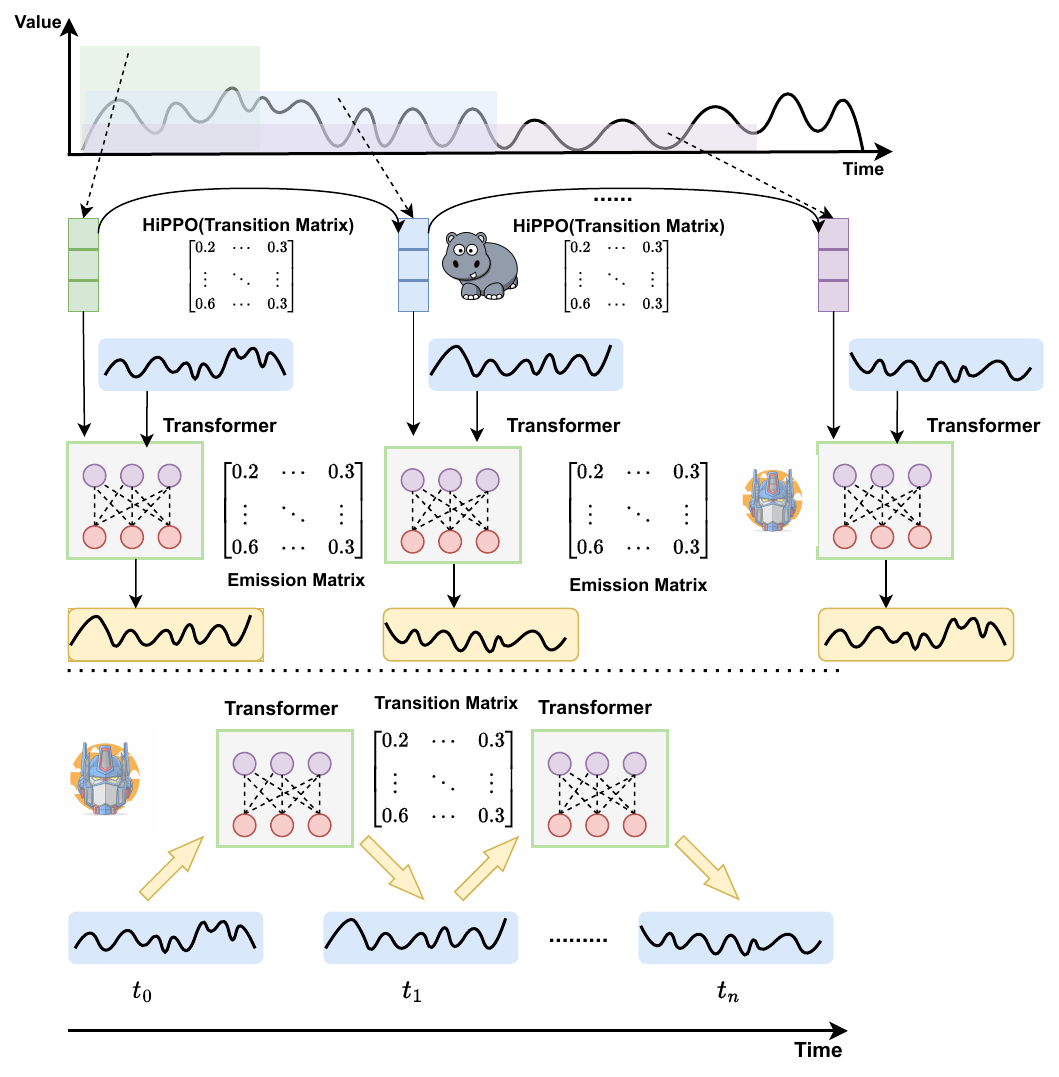}}
    \caption{(a) Structured linear transformation (Right) \textit{vs.} Linear transformation (Left). The temporal constraint of the series is preserved by setting the weights of successor elements to 0, ensuring that these elements do not influence the current element. (b) Markov forecasting process (Bottom) \textit{vs.} Forecasting process with cumulative historical state (Top). A model using only the look-back as input essentially operates as a Markov process, where forecasting is the modeling of the transition matrix. In contrast, our model leverages the cumulative historical state to retain the state of a more complete historical series, with forecasting corresponding to the modeling of the emission matrix.}
    \label{fig1: introduction}
    \vspace{-5mm}
\end{figure}

However, channel-wise Transformers face two main challenges: (1) lacking a mechanism to capture cumulative historical states beyond the look-back window, and (2) using unconstrained linear transformations for temporal feature extraction, which violates fundamental temporal assumptions.

Most current forecasting frameworks rely on a fixed-size historical window, referred to as the look-back window, to predict the next segment of a time series. This approach can be viewed as a first-order Markov process, where the forecasting model approximates the transition matrix. However, this method overlooks the cumulative historical state information accumulated prior to the look-back window, which could enhance model performance if utilized effectively. In terms of feature extraction, channel-wise Transformer employs self-attention to compute correlations among channels, while temporal features are derived through linear transformations and activation functions within the Transformer. Unlike generic sequences, time series have a fundamental temporal constraint: operations on later elements should not influence anterior ones. This assumption is grounded in the sequential nature of time series, where events occurring later cannot retroactively affect earlier events. However, applying unconstrained linear transformations to input or embedded time series violates this assumption, potentially leading to incorrect feature learning and overfitting.

To address these challenges, we employ HiPPO~\cite{gu2020hippo} (High-order Polynomial Projection Operators) to efficiently capture the cumulative historical state. HiPPO recursively embeds long and variable-length time series into a fixed-size state space using orthogonal polynomial bases, providing a simple yet effective memory mechanism that incorporates historical information beyond the look-back window. In this framework, the cumulative historical state functions as the memory state~\cite{chen2023long,hua2019deep}, the HiPPO matrix serves as the transition matrix to model memory updates, and the channel-wise Transformer operates as the emission matrix for forecasting. Fig. \ref{fig1: introduction}(b) highlights the differences between this approach and traditional forecasting methods.

We propose using structured matrices to enforce temporal constraints on linear transformations in channel-wise Transformer. For example, a triangular matrix preserves temporal order by ensuring that elements in the time series embeddings are not influenced by future values, as illustrated in Fig. \ref{fig1: introduction}(a). In this structure, weights assigned to successor elements are set to zero, effectively excluding them from feature computations. Moreover, since 1D convolutions~\cite{li2017classification,kiranyaz20191} inherently respect temporal order, substituting linear transformations in Transformers with 1D convolutions also enforces this constraint. As demonstrated in Chapter \ref{c3}, multi-layer 1D convolutions are mathematically equivalent to a triangular matrix with shared parameters, and the convolution operation can be expressed as a linear transformation using such matrices. This structured design is applied to all linear operations in Transformer, including those in feed-forward layers and the query, key, and value matrices.

Our approach differs from simply extending the fix-size look-back window as input, which fails to capture information beyond the window due to the evolving nature of cumulative history. Moreover, directly using over-long look-back windows can blur the distinction between global features and short-term temporal dependencies~\cite{zeng2023transformers}, making it harder for the model to disentangle these two aspects. In contrast, our method integrates both perspectives: it captures global features through the cumulative historical state while extracting short-term temporal dependencies within the look-back window. Empirical results in Section \ref{c1} demonstrate that these two types of information are decoupled.

The main contributions of this paper are as follows:
\begin{itemize}
    \item HiPPO is introduced to model the cumulative historical state, enhancing the utilization of historical information.
    \item This paper introduces two structured linear transformations, triangular matrices and one-dimensional convolutions, to impose temporal constraints on channel-wise Transformers.
    \item Extensive comparisons and ablations validate the method's effectiveness on real-world datasets.
\end{itemize}

\section{Related Work}
The Transformer~\cite{vaswani2017attention}, a powerful sequence model, has been widely applied in NLP~\cite{wolf2020transformers,kalyan2021ammus,zhao2023transformer}, computer vision~\cite{parvaiz2023vision,vasu2023fastvit,liu2023survey}, and other fields. Leveraging its self-attention mechanism for global sequence modeling, it has become a backbone for many time series forecasting tasks. We summarize efforts to adapt Transformer for these tasks from various perspectives.
\vspace{-5mm}
\subsubsection{Improving Quadratic Computation Cost for Transformer} The vanilla self-attention mechanism has quadratic complexity, making it impractical for long time series. Informer~\cite{zhou2021informer} introduces the ProbSparse self-attention with $O(L \log L)$ complexity and reduces input length via self-attention distillation. FEDformer~\cite{zhou2022fedformer} enhances self-attention with Fourier and Wavelet blocks, achieving linear complexity by selecting a fixed number of Fourier components. Autoformer~\cite{wu2021autoformer} uses series decomposition preprocessing and a deep decomposition architecture to extract predictable components. It replaces point-wise attention with an Auto-Correlation mechanism for series-wise connection, also with $O(L \log L)$ complexity.
\vspace{-5mm}
\subsubsection{Transformer with Patching} PatchTST~\cite{nie2022time} reduces computational complexity by dividing time series into segments as input tokens, which also carry richer semantic information. It employs a channel-independent strategy to simplify training. CARD~\cite{wang2023card} applies self-attention along the time axis for patches and across channels to simultaneously capture temporal and channel features. Pathformer~\cite{chen2024pathformer} integrates multi-scale patch features with dual attention, capturing global correlations and local temporal dependencies.
\vspace{-5mm}
\subsubsection{Channel-wise Transformer} iTransformer~\cite{liu2023itransformer} and SAMformer~\cite{ilbert2024unlocking} apply self-attention to the channels, leveraging its ability to capture inter-channel correlations. It offers a novel perspective for applying Transformer to multivariate time series tasks. Our method also employs channel-wise self-attention but emphasizes timing constraints in feature generation and the use of cumulative historical state.
\vspace{-5mm}
\subsubsection{Cumulative Historical Utilization} SWLHT~\cite{LIU202226} uses short- and long-term memory mechanisms with self-attention to maintain series state information, approximating cumulative historical series. In contrast, our approach employs HiPPO~\cite{gu2020hippo,gu2023mamba} embedding as a standalone module, offering a broader time horizon.

\section{METHOD}
In Multivariate Time Series Forecasting (MTSF), the goal is to predict future time series $\bm{Y} \in \mathbb{R}^{H\times C}$ from a historical multivariate time series (MTS) $\bm{X} \in \mathbb{R}^{L\times C}$, where $H$ is the forecast horizon, $L$ is the look-back window, and $C$ is the number of variables or channels.

\begin{figure}
    \centering
    \includegraphics[scale=0.3]{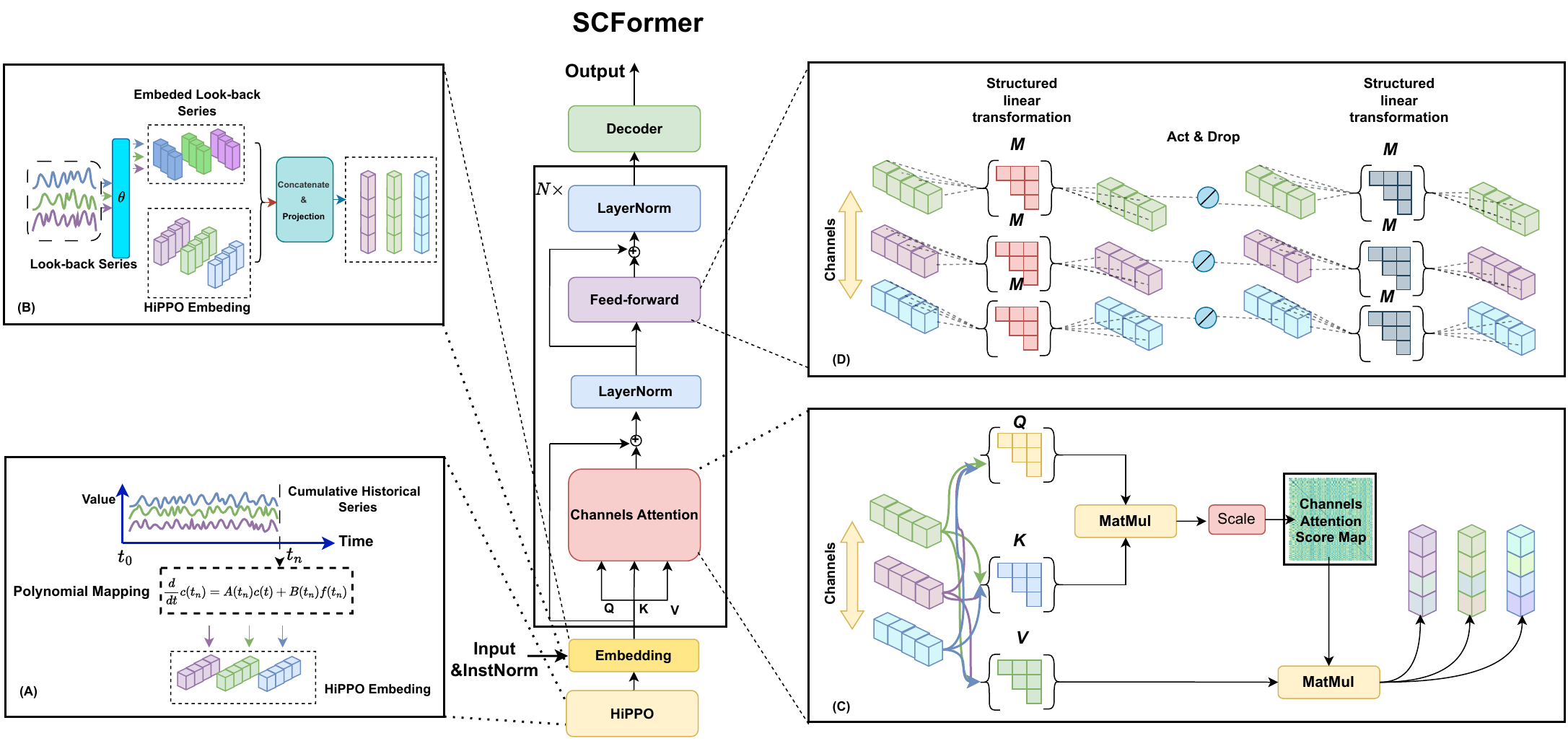}
    \caption{Overall structure of SCFormer. For forecasting at a given moment, the model first computes the cumulative historical state via HiPPO and combines it with the look-back as the final input. Then, temporal constraints are applied to the feature computation through multiple structured linear transformations in the channel-wise Transformer. (A) Cumulative historical state via HiPPO; (B) Embedding; (C) Structured channel-wise self-attention; (D) Structured feed-forward layer.}
    \label{fig:overall architecture}
\end{figure}

Our method consists of two key components: (1) utilizing HiPPO to retain the cumulative historical state and (2) employing structured matrices to enforce temporal constraints on linear transformations in the channel-wise Transformer. The model integrates the look-back window and the cumulative historical state into a unified time series representation and applies a structured channel-wise Transformer to extract temporal and channel correlation features from this time series. SCFormer incorporates a single-layer fully connected network as the decoder and uses Mean Square Error (MSE) as the loss function. The architecture of SCFormer is illustrated in Fig. \ref{fig:overall architecture}.

\begin{figure}[t]
    \centering
    \includegraphics[scale=0.15]{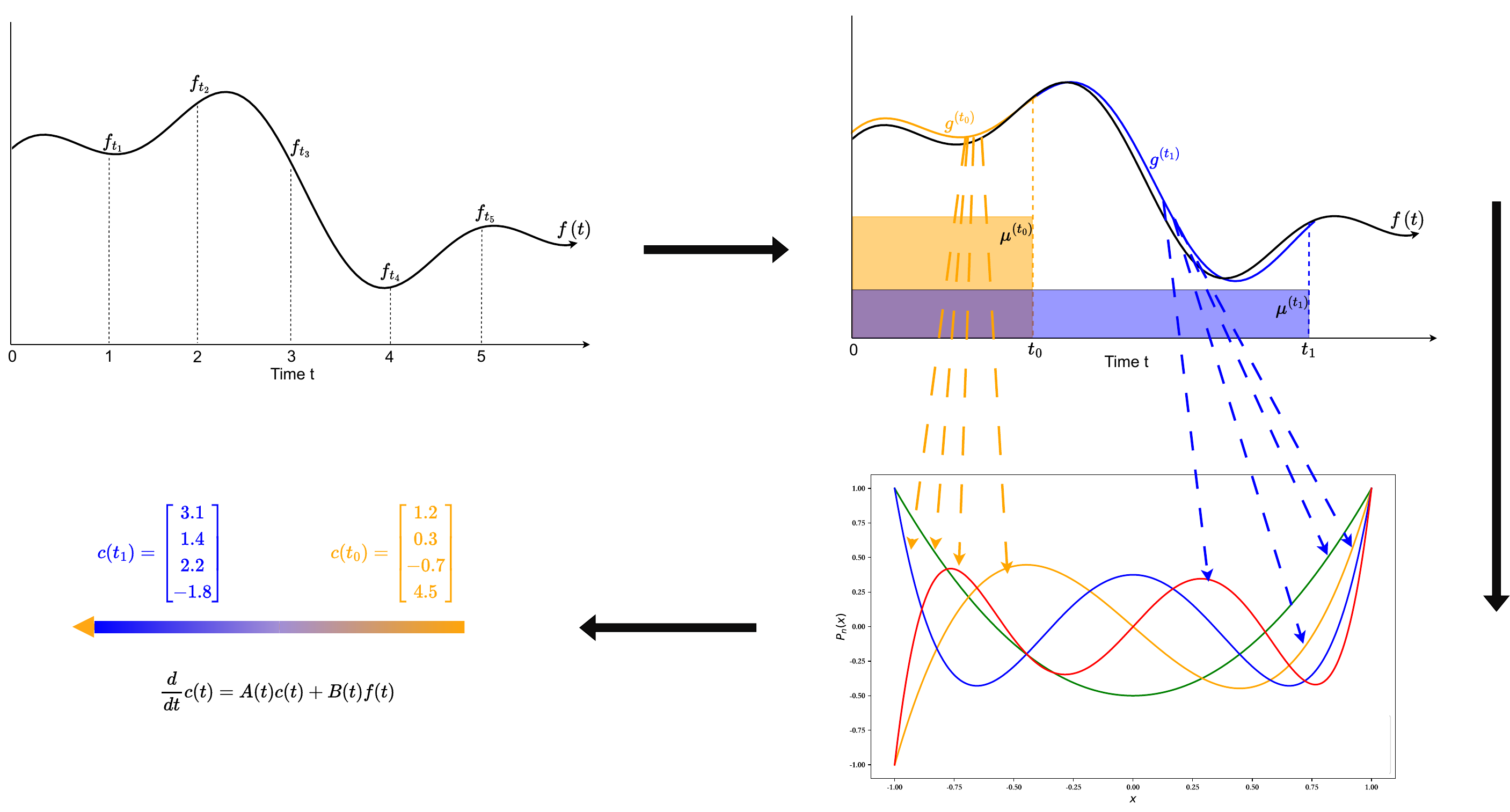}
    \caption{The computation process of HiPPO: The coefficients $c$ are obtained by projecting the sequence $f(t)$ onto an orthogonal polynomial basis under the metric $u$. These coefficients represent the optimal parameters when approximating the sequence $f(t)$ using the orthogonal polynomial basis. HiPPO enables efficient recursive computation through state-space equations.}
    \label{fig:time}
\end{figure}

\subsection{Cumulative Historical State}
The accumulated history includes the entire sequence from the start of the time series up to the current look-back window. As the fixed-size look-back window slides forward, the accumulated history becomes a variable-length series, growing longer over time, which makes it challenging for the model to utilize effectively. To address this, we use HiPPO to compute the cumulative historical state, enabling the model to access richer historical information. HiPPO projects variable-length series onto orthogonal higher-order polynomial bases, embedding the cumulative historical state into a fixed-dimensional space represented by coefficients. This process can be computed efficiently using state-space equations, making it particularly suitable for variable-length sequences. Figure~\ref{fig:time} illustrates the HiPPO computation process. For a time series $\mathbf{x}$, the cumulative historical state $c_{k+1}$ can be computed recursively as follows:

\begin{equation}
\begin{aligned}
    &c_{k+1}=(1-\frac{A}{k})c_k+\frac{1}{k}B\mathbf{x}_{k}, \\
    &A_{nk}=\left\{
    \begin{aligned}
        & (2n+1)^{1/2}(2k+1)^{1/2}\quad if\quad n > k, \\
        & n+1 \quad if\quad n=k,  \\
        & 0 \quad if\quad n < k
    \end{aligned}
    \right. \\
    &B_{n}=(2n+1)^{\frac{1}{2}}
\end{aligned}
\end{equation}
Here, $c_k$ represents the cumulative historical state of $\mathbf{x}_{:\leq k}$. Essentially $c_k$ is the projection coefficient of the history series of $\mathbf{x}_{:\leq k}$ on the orthogonal polynomial basis. $\mathbf{x}_{:\leq k}$ represents the historical series from the beginning timestamp of the $\mathbf{x}$ up to the $k$-th timestamp. $n$ represents the degree of the orthogonal polynomial in HiPPO, which also defines the dimensionality of the cumulative historical state, while $k$ serves as the timestamp indicator.

SCFormer embeds the cumulative historical state and the look-back window into a unified time series. This unified representation encapsulates both global information from the cumulative history and local dependencies from the adjacent window, offering a more comprehensive characterization of temporal patterns. For a look-back window 
$l$ and its corresponding cumulative historical state (HiPPO embedding) $c$, this integration is achieved through concatenation and MLP:
\begin{equation}
    Z = MLP(Concat([MLP(l),c]))
\end{equation}
The operation is both simple and effective: $Z$ incorporates more historical information than $l$ and $c$, which is essential for subsequent feature calculations. It is important to note that temporal constraints do not need to be enforced when computing $Z$, as the cumulative historical state $c$ is not a time series but rather a set of coefficients for polynomial bases. The purpose of $Z$ is to induce a new time series from $l$ and $c$. As long as $Z$ adheres to temporal constraints during subsequent feature calculations, this purpose is satisfied. The structured matrices in SCFormer are specifically designed to ensure this condition is met. From the perspective of a memory mechanism, $c$ functions as the memory state, the HiPPO matrix serves as the state transition matrix, and the channel-wise Transformer models the emission matrix used for forecasting.

\subsection{Triangular Matrix and Temporal Constraint}
\label{TM}
In channel-wise Transformer, the self-attention mechanism involves multiple linear transformations, but these lack temporal constraints. The core issue is that standard matrix multiplication disrupts the series' temporal order, as future elements can influence past ones. For instance, for the $i$-th element $x_i$ in the time series $x$, we calculate its corresponding feature $a_i$ using a linear transformation. According to the matrix multiplication formula:
\begin{equation}
    a_i=\sum_j w_{ij}x_j
\end{equation}
It is evident that all elements in the time series $x$ are involved in the calculation, which is unreasonable. For the set $M=\{x_j, j>i\}$, containing elements that occur after $x_i$, these elements should not influence the generation of $a_i$.

To address this issue, one approach is to set a portion of the matrix elements $W=\{w_{ij}, j>i\}$ to zero. Clearly, this results in a triangular matrix. An upper or lower triangular matrix does not affect temporal constraints, as it merely pertains to whether the growth direction of time is represented using proximal or distal methods. Without loss of generality, this paper adopts an upper triangular matrix as the structured matrix. All linear transformations in the channel-wise Transformer, including the query, key, and value matrices, should follow this structured approach. For the input $\mathbf{Z}\in \mathcal{R}^{d\times C}$, SCFormer applies a channel-wise self-attention mechanism with temporal constraints enforced by structured matrices. Specifically, it calculates the attention scores between channels as follows:

\begin{equation}
\begin{aligned}
        &\mathbf{Q},\mathbf{K},\mathbf{V} = \delta (\mathbf{A}\mathbf{Z}+\mathbf{a}), \delta(\mathbf{B}\mathbf{Z}+\mathbf{b}), \delta(\mathbf{E}\mathbf{Z}+\mathbf{e}) \\
        &s.t \qquad \mathbf{A}_{ij},\mathbf{B}_{ij},\mathbf{C}_{ij}=0, \quad if \quad i > j
\end{aligned}
\end{equation}

\begin{equation}
    attn^i=\frac{\mathbf{Q}^i(\mathbf{K}^i)^T}{\sqrt{d/H}}
\end{equation}
Here, $d$ represents the length of the embedded time series $\mathbf{Z}$, $C$ denotes the number of channels, and $attn^i$ refers to the attention scores of the $i$-th head in the multi-head attention mechanism. $H$ denotes the number of the multi-head. $\mathbf{A}$,$\mathbf{B}$ and $\mathbf{E}$ represent the mapping matrix of query, key and value, and $\mathbf{a}$,$\mathbf{b}$ and $\mathbf{e}$ are the corresponding biases. $\delta$ denotes the Relu activation function.

Subsequently, the output corresponding to each head is obtained using its respective attention scores. Finally, these outputs are concatenated and passed through a structured linear transformation to produce the final output.
\begin{equation}
    \tilde{\mathbf{X}}^i=attn^i\mathbf{V}^i 
\end{equation}

\begin{equation}
    \tilde{\mathbf{X}} = \delta(\mathbf{F}Concat([\tilde{\mathbf{X}}^1, \tilde{\mathbf{X}}^2,...,\tilde{\mathbf{X}}^H])+\mathbf{f}) \qquad s.t \quad \mathbf{F}_{ij}=0, \quad if \quad i > j
\end{equation}

$\mathbf{F}$ represents the weight matrix in the feed-forward layer, and $\mathbf{f}$ is the corresponding bias. It is worth emphasizing that SCFormer captures temporal features through structured linear transformations and activation functions, while the self-attention mechanism is used to compute correlation features between channels. SCFormer is constructed by stacking multiple layers of channel-wise self-attention mechanisms with structured linear transformation.

\subsection{Convolutional Self-attention}
\label{c3}
Another approach to enforcing temporal constraints in the self-attention mechanism is through the use of 1D convolutions. Replacing all linear operations in the Transformer with 1D convolutions ensures that the self-attention mechanism inherits the temporal properties of the convolution. In fact, multi-layer 1D convolutions are mathematically equivalent to a linear transformation implemented using a triangular matrix, offering a more structured approach with shared parameters. For an input series $\mathbf{z}\in \mathcal{R}^{d}$, a convolution with a kernel size of $k$ and stride 1 can be represented as a linear transformation based on a structured matrix $\mathbf{K}$, assuming zero bias for simplicity.

\begin{equation}
\mathbf{K}=
    \begin{bmatrix}
    w_1 & w_2 & \cdots & w_k & \cdots & 0 & 0 &0 \\
    0 & w_1 & \cdots & w_k & \cdots & 0 & 0 & 0 \\
    0 & 0 & w_1 & w_2 & \cdots & w_k & \cdots & 0 \\
    \vdots & \vdots & \vdots & \ddots & \vdots & \vdots & \vdots & \vdots \\
    0 & 0 & 0 & \cdots & 0 & 0 & \cdots & w_1
    \end{bmatrix}
\end{equation}

\begin{equation}
    K * \mathbf{z} = \mathbf{K}\mathbf{z}
\end{equation}

The matrix $\mathbf{K}$ is essentially a Toeplitz matrix. Here, $*$ denotes the convolution operation, and $w_i$ represents the $i$-th weight in the convolution kernel $K$. For multi-layer convolutions, let $\mathbf{K}_i$ be the matrix for each layer. Then, the multi-layer convolutions can be represented as the multiplication of matrices:

\begin{equation}
    \mathcal{F}(\mathbf{z}, k) = (\prod_i \mathbf{K}_i)\mathbf{z}
\end{equation}

Using mathematical induction, it can be shown that the structured form of the matrix $\mathbf{K}_i$ allows the generation of a complete upper triangular matrix with at most $\lceil \frac{d-k}{k-1} \rceil+1$ layers of convolution. This demonstrates that multi-layer 1D convolutions can be implemented as a linear transformation based on an upper triangular matrix with shared weights. The entire convolutional self-attention mechanism can be formalized as follows.

\begin{equation}
    \mathbf{Q},\mathbf{K},\mathbf{V} = \delta(Conv_Q(\mathbf{Z})), \delta(Conv_K(\mathbf{Z})), \delta(Conv_V(\mathbf{Z}))
\end{equation}

\begin{equation}
    attn^i=\frac{\mathbf{Q}^i(\mathbf{K}^i)^T}{\sqrt{d/H}}
\end{equation}

\begin{equation}
    \tilde{\mathbf{X}}^i=attn^i\mathbf{V}^i 
\end{equation}

\begin{equation}
    \tilde{\mathbf{X}} = \delta(Conv_F(Concate([\tilde{\mathbf{X}}^1, \tilde{\mathbf{X}}^2, ...,\tilde{\mathbf{X}}^H])))
\end{equation}
Most of the mathematical symbols are defined in Section~\ref{TM} and will not be repeated here.

\subsection{Instance Normalized and Loss Function}
There is a distribution shift effect in a long time series, which can disturb forecasting performance. To mitigate this problem, the instance normalization technique is proposed~\cite{ulyanov2016instance,kim2022reversible}. It normalizes each look-back series $\mathbf{x}^{(i)}$ to have zero mean and unit standard deviation, and the mean and standard deviation are added back to the final forecast $\hat{\mathbf{Y}}^{(i)}$:

\begin{align}
    &\mathbf{x}^{(i)} = \frac{\mathbf{x}^{(i)} - mean(\mathbf{x}^{(i)})}{stdev(\mathbf{x}^{(i)})} \notag \\ 
    &\hat{\mathbf{Y}}^{(i)} = [\hat{\mathbf{Y}}^{(i)} + mean(\mathbf{x}^{(i)})]*stdev(\mathbf{x}^{(i)})
\end{align}

We use a simple fully connected network as the decoder. Following most previous works, we use the Mean Squared Error (MSE) Loss, which measures the average squared difference between the predicted values and the ground truth. The definition of the loss function is as follows:
\begin{equation}
    \mathcal{L}(Y, \hat{Y}) = \frac{1}{|Y|} \sum_{i=1}^{|Y|} \left( y_{(i)} - \hat{y}_{(i)} \right)^2
\end{equation}
where $\hat{Y}$ is the predicted values and $Y$ is the ground truth.

\section{Experiments}
\subsection{Datasets and implementation}

\subsubsection{Datasets}
We evaluate our method on several widely used datasets. ETT~\cite{zhou2021informer} (subsets: ETTh1, ETTh2, ETTm1, ETTm2), we report average performance on them; Electricity (ECL)\footnote{\url{https://archive.ics.uci.edu/ml/datasets/ElectricityLoadDiagrams20112014}}; Traffic\footnote{\url{http://pems.dot.ca.gov}}; Weather\footnote{\url{https://www.bgc-jena.mpg.de/wetter/}}; Exchange~\cite{lai2018modeling}; The PEMS (subsets PEMS04 and PEMS07)~\cite{liu2022scinet}; Solar-Energy~\cite{lai2018modeling}. Most datasets are split in a 7:1:2 ratio for training, validation, and testing. Details are in Table \ref{tab: data descriptions}.

\begin{table}[]
    \centering
    \scalebox{0.85}{
    \begin{tabular}{c|c|c|c}
    \toprule
    Dataset & Dim & Size & Frequency \\
    \midrule
    ETTh1,ETTh2 & 7 & 8545, 2881, 2881 & 1h \\
    ETTm1, ETTm2 & 7 & 34465, 11521, 11521 & 15m \\
    Exchage & 8 & 5120, 665, 1422 & 1d \\
    Weather & 21 & 36792, 5271, 10540 & 10m \\
    ECL & 321 & 18317, 2633, 5261 & 1h \\
    Traffic & 862 & 12185, 1757, 3509 & 1h \\
    Solar-Energy & 137 & 36601, 5161, 10417 & 10m \\
    PEMS04 & 307 & 10172, 3375, 281 & 5m \\
    PEMS07 & 883 & 16911, 5622, 468 & 5m \\
    \bottomrule
    \end{tabular}
    }
    \caption{Details of benchmark datasets.}
    \label{tab: data descriptions}
    \vspace{-10mm}
\end{table}

\subsubsection{Implementation} Since iTransformer~\cite{liu2023itransformer} is also a channel-wise Transformer, we use the same configuration for Transformer-related hyperparameters as those used in iTransformer, with the HiPPO order set to 512. For the convolutional self-attention mechanism, we use 3 convolutional layers with a kernel size of 32 and a stride of 1. The evaluation metrics include mean squared error (MSE) and mean absolute error (MAE). Experiments are conducted on an NVIDIA V100 GPU with 32GB of memory, using PyTorch 1.13.1.

\subsection{Baselines}
Nine popular methods are used as baselines, including four Transformer-based methods: (1) iTransformer~\cite{liu2023itransformer}, (2) FEDformer~\cite{zhou2022fedformer}, (3) PatchTST~\cite{nie2022time}, and (4) Crossformer~\cite{zhang2022crossformer}; three MLP-based methods: (5) TiDE~\cite{das2023long}, (6) RLinear~\cite{li2023revisiting}, and (7) DLinear~\cite{zeng2023transformers}; and two CNN-based methods: (8) TimesNet~\cite{wu2022timesnet} and (9) SCINet~\cite{liu2022scinet}. The baseline results are all taken from those reported in their respective papers. All methods, including ours, use a fixed look-back size of 96 and predict time horizons of 96, 192, 336, and 720.

\subsection{Experimental Results}
\renewcommand{\dblfloatpagefraction}{.8}
\begin{table}[t]
    \centering
    \scalebox{0.55}{
    \begin{tabular}{c|c|cc|cc|cc|cc|cc|cc|cc|cc|cc|cc|cc}
    \toprule
          \multicolumn{2}{c|}{Models} & \multicolumn{2}{c|}{\makecell[c]{SCFormer\\ \textit{conv}}} & \multicolumn{2}{c|}{\makecell[c]{SCFormer\\ \textit{triangular}}} & \multicolumn{2}{c|}{\makecell[c]{iTransformer}} & \multicolumn{2}{c|}{\makecell[c]{RLinear}} & \multicolumn{2}{c|}{\makecell[c]{PatchTST}} & \multicolumn{2}{c|}{\makecell[c]{Crossformer}} & \multicolumn{2}{c|}{\makecell[c]{TiDE}} & \multicolumn{2}{c|}{\makecell[c]{TimesNet}} & \multicolumn{2}{c|}{\makecell[c]{DLinear}} & \multicolumn{2}{c|}{\makecell[c]{SCINet}} & \multicolumn{2}{c}{\makecell[c]{FEDformer}} \\
    \midrule
    Dateset & H & MSE & MAE & MSE & MAE & MSE & MAE & MSE & MAE & MSE & MAE & MSE & MAE & MSE & MAE & MSE & MAE &  MSE & MAE & MSE & MAE & MSE & MAE\\
    \midrule
    \multirow{5}*{ETT} & 96 & \underline{0.295} & 0.344 & \textbf{0.291} & \textbf{0.338} & 0.299 & 0.346 & 0.302 & \underline{0.343} & 0.305 & 0.348 & 0.464 & 0.456 & 0.362 & 0.399 & 0.312 & 0.354 & 0.314 & 0.362 & 0.516 & 0.508 & 0.329 & 0.380 \\
    ~ & 192 & \underline{0.354} & 0.380 & \textbf{0.350} & \textbf{0.375} & 0.362 & 0.384 & 0.362 & \underline{0.377} & 0.364 & 0.383 & 0.553 & 0.518 & 0.435 & 0.442 & 0.365 & 0.384 & 0.394 & 0.414 & 0.604 & 0.553 & 0.386 & 0.414 \\
    ~ & 336 & \textbf{0.399} & \underline{0.409} & \underline{0.401} & 0.409 & 0.413 & 0.414 & 0.406 & \textbf{0.407} & 0.407 & 0.413 & 0.685 & 0.583 & 0.503 & 0.483 & 0.418 & 0.420 & 0.464 & 0.460 & 0.726 & 0.619 & 0.431 & 0.444 \\
    ~ & 720 & \textbf{0.444} & \underline{0.441} & 0.460 & 0.454 & 0.458 & 0.450 & 0.448 & \textbf{0.439} & \underline{0.446} & 0.443 & 1.038 & 0.753 & 0.628 & 0.555 & 0.467 & 0.455 & 0.594 & 0.537 & 0.910 & 0.705 & 0.483 & 0.471 \\
    ~ & Avg & \textbf{0.373} & \underline{0.393} & \underline{0.375} & 0.394 & 0.383 & 0.399 & 0.380 & \textbf{0.392} & 0.381 & 0.397 & 0.685 & 0.578 & 0.482 & 0.470 & 0.391 & 0.404 & 0.442 & 0.444 & 0.689 & 0.597 & 0.408 & 0.428 \\
    \midrule
    \multirow{5}*{PEMS} & 96 & 0.078 & 0.190 & \textbf{0.067} & \textbf{0.167} & 0.072 & 0.174 & 0.128 & 0.243 & 0.100 & 0.215 & 0.096 & 0.209 & 0.196 & 0.322 & 0.084 & 0.188 & 0.131 & 0.257 & \underline{0.070} & \underline{0.174} & 0.123 & 0.243\\
    ~ & 192 & \underline{0.091} & 0.203 & \textbf{0.078} & \textbf{0.181} & 0.091 & \underline{0.197} & 0.250 & 0.344 & 0.151 & 0.268 & 0.135 & 0.251 & 0.281 & 0.390 & 0.102 & 0.209 & 0.217 & 0.334 & 0.101 & 0.209 & 0.151 & 0.268 \\
    ~ & 336 & \underline{0.110} & \underline{0.220} & \textbf{0.089} & \textbf{0.193} & 0.115 & 0.224 & 0.567 & 0.542 & 0.241 & 0.339 & 0.258 & 0.347 & 0.427 & 0.486 & 0.135 & 0.244 & 0.376 & 0.447 & 0.124 & 0.224 & 0.217 & 0.328 \\
    ~ & 720 & 0.137 & 0.242 & \textbf{0.104} & \textbf{0.207} & 0.144 & 0.253 & 1.116 & 0.807 & 0.318 & 0.396 & 0.399 & 0.449 & 0.560 & 0.554 & 0.185 & 0.291 & 0.523 & 0.528 & \underline{0.127} & \underline{0.230} & 0.301 & 0.401 \\
    ~ & Avg & \underline{0.104} & 0.213 & \textbf{0.084} & 0.187 & 0.105 & 0.212 & 0.515 & 0.484 & 0.202 & 0.305 & 0.222 & 0.314 & 0.366 & 0.438 & 0.126 & 0.233 & 0.311 & 0.391 & 0.105 & \underline{0.209} & 0.198 & 0.310 \\
    \midrule
    \multirow{5}*{Solar Energy} & 96 & \underline{0.199} & 0.251 & \textbf{0.193} & \textbf{0.231} & 0.203 & \underline{0.237} & 0.322 & 0.339 & 0.234 & 0.286 & 0.310 & 0.331 & 0.312 & 0.399 & 0.250 & 0.292 & 0.290 & 0.378 & 0.237 & 0.344 & 0.242 & 0.342 \\
    ~ & 192 & \underline{0.229} & 0.271 & \textbf{0.224} & \textbf{0.259} & 0.233 & \underline{0.261} & 0.359 & 0.356 & 0.267 & 0.310 & 0.734 & 0.725 & 0.339 & 0.416 & 0.296 & 0.318 & 0.320 & 0.398 & 0.280 & 0.380 & 0.285 & 0.380 \\
    ~ & 336 & \underline{0.248} & 0.287 & \textbf{0.242} & \underline{0.274} & 0.248 & \textbf{0.273} & 0.397 & 0.369 & 0.290 & 0.315 & 0.750 & 0.735 & 0.368 & 0.430 & 0.319 & 0.330 & 0.353 & 0.415 & 0.304 & 0.389 & 0.282 & 0.376 \\
    ~ & 720 & 0.252 & 0.287 & \underline{0.251} & \underline{0.281} & \textbf{0.249} & \textbf{0.275} & 0.397 & 0.356 & 0.289 & 0.317 & 0.769 & 0.765 & 0.370 & 0.425 & 0.338 & 0.337 & 0.356 & 0.413 & 0.308 & 0.388 & 0.357 & 0.427 \\
    ~ & Avg & \underline{0.232} & 0.274 & \textbf{0.227} & \textbf{0.261} & 0.233 & \underline{0.262} & 0.369 & 0.356 & 0.270 & 0.307 & 0.641 & 0.639 & 0.347 & 0.417 & 0.301 & 0.319 & 0.330 & 0.401 & 0.282 & 0.375 & 0.291 & 0.381 \\
    \midrule
    \multirow{5}*{ECL} & 96 & \underline{0.134} & \underline{0.233} & \textbf{0.129} & \textbf{0.228} & 0.148 & 0.240 & 0.201 & 0.281 & 0.195 & 0.285 & 0.219 & 0.314 & 0.237 & 0.329 & 0.168 & 0.272 & 0.197 & 0.282 & 0.247 & 0.345 & 0.193 & 0.308 \\
    ~ & 192 & \underline{0.150} & \underline{0.247} & \textbf{0.147} & \textbf{0.245} & 0.162 & 0.253 & 0.201 & 0.283 & 0.199 & 0.289 & 0.231 & 0.322 & 0.236 & 0.330 & 0.184 & 0.289 & 0.196 & 0.285 & 0.257 & 0.355 & 0.201 & 0.315 \\
    ~ & 336 & \underline{0.167} & \underline{0.264} & \textbf{0.160} & \textbf{0.260} & 0.178 & 0.269 & 0.215 & 0.298 & 0.215 & 0.305 & 0.246 & 0.337 & 0.249 & 0.344 & 0.198 & 0.300 & 0.209 & 0.301 & 0.269 & 0.369 & 0.214 & 0.329 \\
    ~ & 720 & \underline{0.195} & \underline{0.292} & \textbf{0.191} & \textbf{0.286} & 0.225 & 0.317 & 0.257 & 0.331 & 0.256 & 0.337 & 0.280 & 0.363 & 0.284 & 0.373 & 0.220 & 0.320 & 0.245 & 0.333 & 0.299 & 0.390 & 0.246 & 0.355 \\
    ~ & Avg & \underline{0.161} & \underline{0.259} & \textbf{0.156} & \textbf{0.254} & 0.178 & 0.270 & 0.219 & 0.298 & 0.216 & 0.304 & 0.244 & 0.334 & 0.251 & 0.344 & 0.192 & 0.295 & 0.212 & 0.300 & 0.268 & 0.365 & 0.214 & 0.327 \\
    \midrule
    \multirow{5}*{Exchange} & 96 & \textbf{0.085} & \underline{0.208} & \underline{0.086} & 0.209 & 0.086 & \textbf{0.206} & 0.093 & 0.217 & 0.088 & 0.205 & 0.256 & 0.367 & 0.094 & 0.218 & 0.107 & 0.234 & 0.088 & 0.218 & 0.267 & 0.396 & 0.148 & 0.278 \\
    ~ & 192 & \textbf{0.171} & \underline{0.298} & \textbf{0.171} & \textbf{0.295} & 0.177 & 0.299 & 0.184 & 0.307 & 0.176 & 0.299 & 0.470 & 0.509 & 0.184 & 0.307 & 0.226 & 0.344 & 0.176 & 0.315 & 0.351 & 0.459 & 0.271 & 0.315 \\
    ~ & 336 & 0.324 & 0.412 & \textbf{0.296} & \textbf{0.395} & 0.331 & 0.417 & 0.351 & 0.432 & \underline{0.301} & \underline{0.397} & 1.268 & 0.883 & 0.349 & 0.431 & 0.367 & 0.448 & 0.313 & 0.427 & 1.324 & 0.853 & 0.460 & 0.427 \\
    ~ & 720 & \underline{0.682} & \underline{0.623} & \textbf{0.645} & \textbf{0.612} & 0.847 & 0.691 & 0.886 & 0.714 & 0.901 & 0.714 & 1.767 & 1.068 & 0.852 & 0.698 & 0.964 & 0.746 & 0.839 & 0.695 & 1.058 & 0.797 & 1.195 & 0.695 \\
    ~ & Avg & \underline{0.315} & \underline{0.385} & \textbf{0.299} & \textbf{0.377} & 0.360 & 0.403 & 0.378 & 0.417 & 0.367 & 0.404 & 0.940 & 0.707 & 0.370 & 0.413 & 0.416 & 0.443 & 0.354 & 0.414 & 0.750 & 0.626 & 0.519 & 0.429 \\
    \midrule
    \multirow{5}*{Weather} & 96 & 0.163 & \underline{0.213} & \textbf{0.156} & \textbf{0.205} & 0.174 & 0.214 & 0.192 & 0.232 & 0.177 & 0.218 & \underline{0.158} & 0.230 & 0.202 & 0.261 & 0.172 & 0.220 & 0.196 & 0.255 & 0.221 & 0.306 & 0.217 & 0.296 \\
    ~ & 192 & \underline{0.209} & \textbf{0.253} & 0.212 & \underline{0.254} & 0.221 & 0.254 & 0.240 & 0.271 & 0.225 & 0.259 & \textbf{0.206} & 0.277 & 0.242 & 0.298 & 0.219 & 0.261 & 0.237 & 0.296 & 0.261 & 0.340 & 0.276 & 0.336 \\
    ~ & 336 & \textbf{0.259} & \textbf{0.292} & \underline{0.261} & \underline{0.293} & 0.278 & 0.296 & 0.292 & 0.307 & 0.278 & 0.297 & 0.272 & 0.335 & 0.287 & 0.335 & 0.280 & 0.306 & 0.283 & 0.335 & 0.309 & 0.378 & 0.339 & 0.380 \\
    ~ & 720 & \underline{0.322} & \underline{0.338} & \textbf{0.313} & \textbf{0.334} & 0.358 & 0.349 & 0.364 & 0.353 & 0.354 & 0.348 & 0.398 & 0.418 & 0.351 & 0.386 & 0.365 & 0.359 & 0.345 & 0.381 & 0.377 & 0.427 & 0.403 & 0.428 \\
    ~ & Avg & \underline{0.238} & \underline{0.274} & \textbf{0.235} & \textbf{0.271} & 0.258 & 0.279 & 0.272 & 0.291 & 0.259 & 0.281 & 0.259 & 0.315 & 0.271 & 0.320 & 0.259 & 0.287 & 0.265 & 0.317 & 0.292 & 0.363 & 0.309 & 0.360 \\
    \midrule
    \multirow{5}*{Traffic} & 96 & \underline{0.408} & 0.296 & 0.448 & 0.333 & \textbf{0.395} & \textbf{0.268} & 0.649 & 0.389 & 0.544 & 0.359 & 0.522 & \underline{0.290} & 0.805 & 0.493 & 0.593 & 0.321 & 0.650 & 0.396 & 0.788 & 0.499 & 0.587 & 0.366 \\
    ~ & 192 & \underline{0.431} & 0.301 & 0.44 & 0.314 & \textbf{0.417} & \textbf{0.276} & 0.601 & 0.366 & 0.540 & 0.354 & 0.530 & \underline{0.293} & 0.756 & 0.474 & 0.617 & 0.336 & 0.598 & 0.370 & 0.789 & 0.505 & 0.604 & 0.373 \\
    ~ & 336 & \underline{0.451} & 0.309 & 0.521 & 0.360 & \textbf{0.433} & 0.283 & 0.609 & 0.369 & 0.551 & 0.358 & 0.558 & \underline{0.305} & 0.762 & 0.477 & 0.629 & 0.336 & 0.605 & 0.373 & 0.797 & 0.508 & 0.621 & 0.383 \\
    ~ & 720 & \underline{0.491} & \underline{0.319} & 0.630 & 0.431 & \textbf{0.467} & \textbf{0.302} & 0.647 & 0.387 & 0.586 & 0.375 & 0.589 & 0.328 & 0.719 & 0.449 & 0.640 & 0.350 & 0.645 & 0.394 & 0.841 & 0.523 & 0.626 & 0.382 \\
    ~ & Avg & \underline{0.445} & 0.306 & 0.509 & 0.359 & \textbf{0.428} & \textbf{0.282} & 0.626 & 0.378 & 0.555 & 0.362 & 0.550 & \underline{0.304} & 0.760 & 0.473 & 0.620 & 0.336 & 0.625 & 0.383 & 0.804 & 0.509 & 0.610 & 0.376 \\
    \bottomrule
    \end{tabular}}
    \caption{The main experimental results of the comparison baselines are presented. SCFormer-\textit{conv} refers to the implementation of temporal constraints using 1D convolutions, while SCFormer-\textit{triangular} refers to the implementation of temporal constraints using triangular matrices. Optimal results are highlighted in bold, and suboptimal results are underlined.}
    \label{tab:main_result}
    \vspace{-10mm}
\end{table}

\begin{table}[htbp]
    \centering
    \begin{minipage}[b]{0.32\textwidth}
        \centering
        \scalebox{0.45}{
    \begin{tabular}{c|c|cc|cc|cc}
    \toprule
          \multicolumn{2}{c|}{Models} & \multicolumn{2}{c|}{\makecell[c]{SCFormer\\ \textit{conv}}} & \multicolumn{2}{c|}{\makecell[c]{SCFormer\\ \textit{triangular}}} & \multicolumn{2}{c}{\makecell[c]{Transformer\\HiPPO}} \\
    \midrule
    Dateset & H & MSE & MAE & MSE & MAE & MSE & MAE \\
    \midrule
    \multirow{4}*{ETTm1} & 96 & 0.322 & 0.362 & 0.318 & \textbf{0.354} & \textbf{0.315} & 0.356 \\
    ~ & 192 & \textbf{0.362} & 0.383 & 0.364 & \textbf{0.382} & 0.370 & 0.387 \\
    ~ & 336 & \textbf{0.394} & \textbf{0.404} & 0.398 & 0.406 & 0.402 & 0.410 \\
    ~ & 720 & \textbf{0.460} & \textbf{0.441} & 0.471 & 0.449 & 0.468 & 0.450 \\
    \midrule
    \multirow{4}*{ETTm2} & 96 & 0.172 & 0.261 & \textbf{0.171} & \textbf{0.256} & 0.175 & 0.261 \\
    ~ & 192 & 0.241 & 0.308 & \textbf{0.232} & \textbf{0.301} & 0.245 & 0.310 \\
    ~ & 336 & \textbf{0.309} & \textbf{0.351} & 0.325 & 0.361 & 0.333 & 0.365 \\
    ~ & 720 & \textbf{0.408} & \textbf{0.406} & 0.454 & 0.439 & 0.445 & 0.433 \\
    \midrule
    \multirow{4}*{ETTh1} & 96 & 0.384 & 0.401 & \textbf{0.374} & \textbf{0.394} & 0.377 & 0.398 \\
    ~ & 192 & 0.434 & 0.430 & \textbf{0.424} & \textbf{0.423} & 0.425 & 0.427 \\
    ~ & 336 & 0.476 & 0.451 & \textbf{0.462} & \textbf{0.444} & 0.471 & 0.455 \\
    ~ & 720 & \textbf{0.483} & \textbf{0.474} & 0.489 & 0.487 & 0.494 & 0.491 \\
    \midrule
    \multirow{4}*{ETTh2} & 96 & 0.302 & 0.352 & \textbf{0.301} & \textbf{0.348} & 0.312 & 0.356 \\
    ~ & 192 & 0.381 & 0.399 & 0.380 & \textbf{0.395} & \textbf{0.379} & 0.399 \\
    ~ & 336 & 0.419 & 0.431 & 0.419 & 0.428 & \textbf{0.416} & \textbf{0.427} \\
    ~ & 720 & \textbf{0.426} & \textbf{0.443} & 0.427 & 0.443 & 0.436 & 0.446 \\
    \bottomrule
    
    \end{tabular}
        }
        \text{(a)}
    \end{minipage}%
    \hfill
    \begin{minipage}[b]{0.32\textwidth}
        \centering
        \scalebox{0.45}{
    \begin{tabular}{c|c|cc|cc}
    \toprule
          \multicolumn{2}{c|}{Models} & \multicolumn{2}{c|}{\makecell[c]{SCFormer\\ \textit{triangular}}} & \multicolumn{2}{c}{\makecell[c]{SCFormer\\ \textit{triangular/wo-HiPPO}}} \\
    \midrule
    Dateset & H & MSE & MAE & MSE & MAE \\
    \midrule
    \multirow{5}*{ECL} & 96 & \textbf{0.129} & \textbf{0.228} & 0.149 & 0.241 \\
    ~ & 192 & \textbf{0.147} & \textbf{0.245} & 0.163 & 0.254 \\
    ~ & 336 & \textbf{0.160} & \textbf{0.260} & 0.179 & 0.270 \\
    ~ & 720 & \textbf{0.191} & \textbf{0.286} & 0.213 & 0.299 \\
    ~ & Avg & \textbf{0.156} & \textbf{0.254} & 0.176 & 0.266 \\
    \midrule
    \multirow{5}*{Weather} & 96 & \textbf{0.156} & \textbf{0.205} & 0.176 & 0.217 \\
    ~ & 192 & \textbf{0.212} & \textbf{0.254} & 0.225 & 0.260 \\
    ~ & 336 & \textbf{0.261} & \textbf{0.293} & 0.282 & 0.301 \\
    ~ & 720 & \textbf{0.313} & \textbf{0.334} & 0.356 & 0.350 \\
    ~ & Avg & \textbf{0.235} & \textbf{0.271} & 0.259 & 0.282 \\
    \midrule
    \multirow{5}*{Solar Energy} & 96 & \textbf{0.193} & \textbf{0.231} & 0.203 & 0.238 \\
    ~ & 192 & \textbf{0.224} & \textbf{0.259}  & 0.238 & 0.265 \\
    ~ & 336 & \textbf{0.242} & \textbf{0.274} & 0.249 & 0.275 \\
    ~ & 720 & \textbf{0.251} & 0.281 & 0.251 & \textbf{0.278} \\
    ~ & Avg & \textbf{0.227} & \textbf{0.261} & 0.235 & 0.264 \\
    \bottomrule
    
    \end{tabular}
        }
        \text{(b)}
    \end{minipage}%
    \hfill
    \begin{minipage}[b]{0.32\textwidth}
        \centering
        \scalebox{0.45}{
    \begin{tabular}{c|c|cc|cc}
    \toprule
          \multicolumn{2}{c|}{Models} & \multicolumn{2}{c|}{\makecell[c]{SCFormer\\ \textit{triangular}}} & \multicolumn{2}{c}{\makecell[c]{SCFormer\\ \textit{triangular/wo-look-back}}} \\
    \midrule
    Dateset & H & MSE & MAE & MSE & MAE \\
    \midrule
    \multirow{5}*{ECL} & 96 & \textbf{0.129} & \textbf{0.228} & 0.137 & 0.242 \\
    ~ & 192 & \textbf{0.147} & \textbf{0.245} & 0.154 & 0.260 \\
    ~ & 336 & \textbf{0.160} & \textbf{0.260} & 0.174 & 0.286 \\
    ~ & 720 & \textbf{0.191} & \textbf{0.286} & 0.203 & 0.314 \\
    ~ & Avg & \textbf{0.156} & \textbf{0.254} & 0.167 & 0.275 \\
    \midrule
    \multirow{5}*{Traffic} & 96 & \textbf{0.448} & \textbf{0.333} & 0.676 & 0.452 \\
    ~ & 192 & \textbf{0.440} & \textbf{0.314} & 0.705 & 0.452 \\
    ~ & 336 & \textbf{0.521} & \textbf{0.360} & 0.794 & 0.479 \\
    ~ & 720 & \textbf{0.630} & \textbf{0.431} & 0.852 & 0.501 \\
    ~ & Avg & \textbf{0.509} & \textbf{0.359} & 0.756 & 0.471 \\
    \midrule
    \multirow{5}*{Solar Energy} & 96 & \textbf{0.193} & \textbf{0.231} & 0.220 & 0.286 \\
    ~ & 192 & \textbf{0.224} & \textbf{0.259}  & 0.233 & 0.279 \\
    ~ & 336 & \textbf{0.242} & \textbf{0.274} & 0.250 & 0.291 \\
    ~ & 720 & \textbf{0.251} & \textbf{0.281} & 0.262 & 0.298 \\
    ~ & Avg & \textbf{0.227} & \textbf{0.261} & 0.241 & 0.288 \\
    \bottomrule
    
    \end{tabular}
        }
        \text{(c)}
    \end{minipage}
    \caption{(a) The ablation experimental results for temporal constraints using structured matrices on the \textit{ETT} dataset. Transformer-HiPPO refers to the channel-wise Transformer equipped with HiPPO. (b) The ablation experimental results without HiPPO. SCFormer-\textit{triangular/wo-HiPPO} represents SCFormer-\textit{triangular} without the cumulative historical state via HiPPO embedding. (c) The ablation experimental results without the look-back window. SCFormer-\textit{triangular/wo-look-back} refers to SCFormer-\textit{triangular} without the look-back window.}
    \label{tab:all_ablation}
    \vspace{-10mm}
\end{table}

Table \ref{tab:main_result} compares the proposed SCFormer-\textit{conv} and SCFormer-\textit{triangular} with baseline methods. SCFormer-\textit{conv} replaces all linear transformations with 1D convolutions, while SCFormer-\textit{triangular} employs structured triangular matrices for linear mappings. Both approaches significantly improve forecasting performance, with SCFormer-\textit{triangular} achieving superior results. For example, SCFormer-\textit{triangular} achieves an average MSE improvement of 12.3\% over the channel-wise state-of-the-art model iTransformer~\cite{liu2023itransformer} on the ECL dataset, 16.9\% on the Exchange dataset, and 8.9\% on the Weather dataset. For SCFormer-\textit{conv}, it achieves an average MSE improvement of 2.6\% on the ETT dataset and 7.7\% on the Weather dataset. Considering the parameter size of SCFormer-\textit{conv}, this performance is quite competitive. It is easy to observe that SCFormer-\textit{triangular}, by using a triangular matrix structure, reduces the model's parameters scale by approximately 50\% compared to the vanilla Transformer. On the other hand, SCFormer-\textit{conv}, which replaces the matrix with 1D convolution kernels, reduces the parameter scale to about 10\% of that in vanilla Transformer. Thus, SCFormer exhibits very high parameter efficiency. An exception is observed on the \textit{traffic} dataset in Table \ref{tab:main_result}; however, SCFormer-\textit{conv} still achieves suboptimal performance. The results demonstrate that most datasets benefit from our method, emphasizing its general effectiveness.

\subsection{Ablation Study}
\subsubsection{Temporal Constraints}
To explore the role of temporal constraints, we compare SCFormer with Transformer-HiPPO, which removes the temporal constraint but retains the HiPPO embedding. As shown in Table \ref{tab:all_ablation}(a), SCFormer achieves better forecasting performance in most circumstances. This suggests that temporal constraints help mitigate overfitting, leading to lower prediction error.

\subsubsection{Cumulative Historical State}
The cumulative historical state maintains the long-term state of a historical series by projecting it into an orthogonal polynomial space using HiPPO. To evaluate its impact on forecasting performance, we remove the HiPPO embedding from the model. As shown in Table \ref{tab:all_ablation}(b), the model's performance significantly declines, demonstrating the effect of the cumulative historical state.

\vspace{-5mm}
\subsubsection{Look-back Window}
To assess the necessity of the look-back window, we remove it and use only the cumulative historical state generated by HiPPO for forecasting. As shown in Table \ref{tab:all_ablation}(c), the model's performance significantly drops without the look-back window. This is expected, as HiPPO represents the overall state of the time series, not direct information in the real number domain. This confirms that the cumulative historical state and look-back window provide complementary features.

\begin{figure}[t]
	\centering
	\subfigure[\scalebox{0.8}{}] {\includegraphics[width=.25\textwidth]{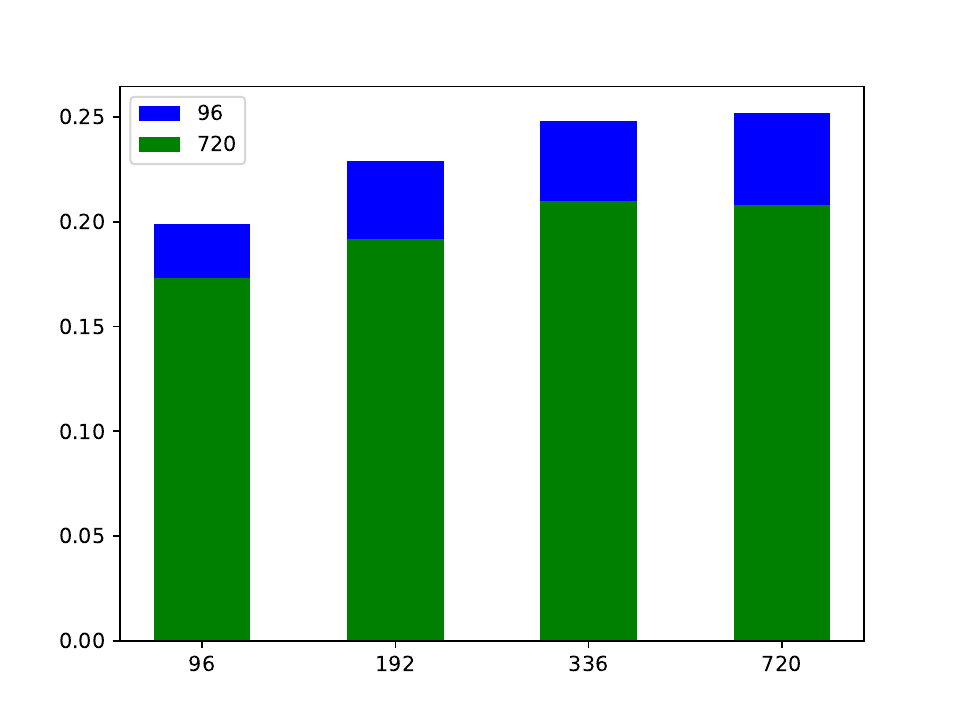}}
	\subfigure[\scalebox{0.8}{}] {\includegraphics[width=.25\textwidth]{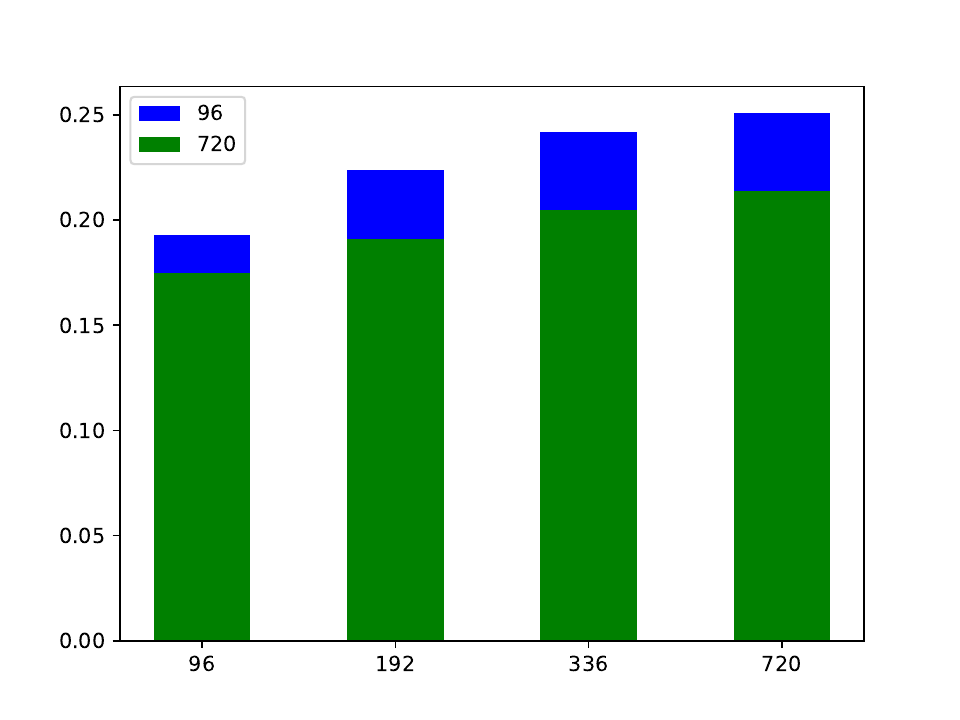}}
	\subfigure[\scalebox{0.8}{}] {\includegraphics[width=.25\textwidth]{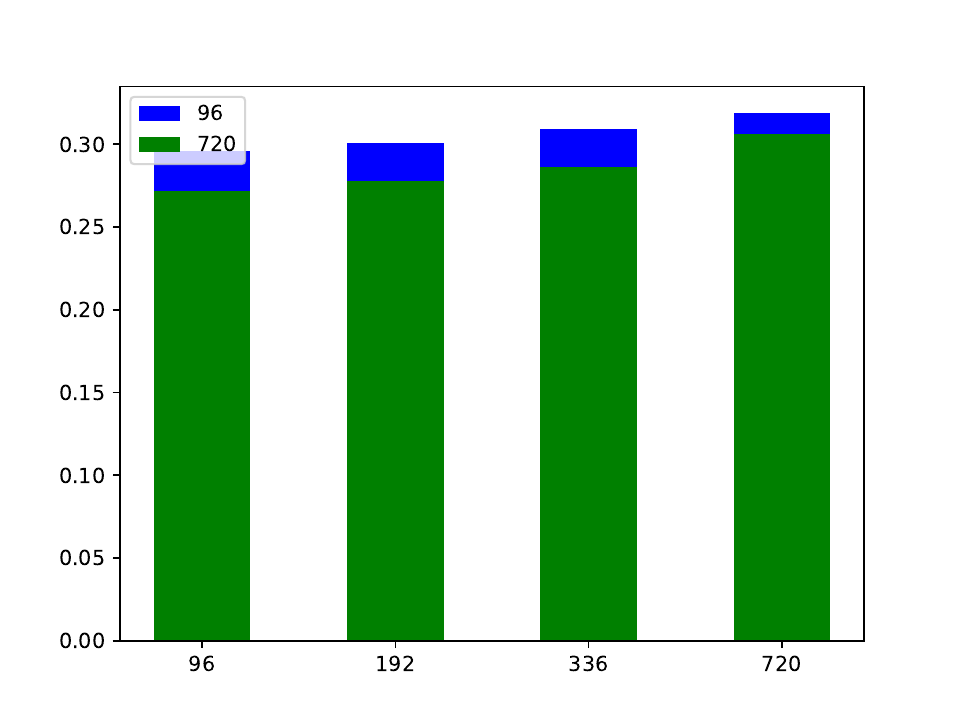}}
    \subfigure[\scalebox{0.8}{}] {\includegraphics[width=.25\textwidth]{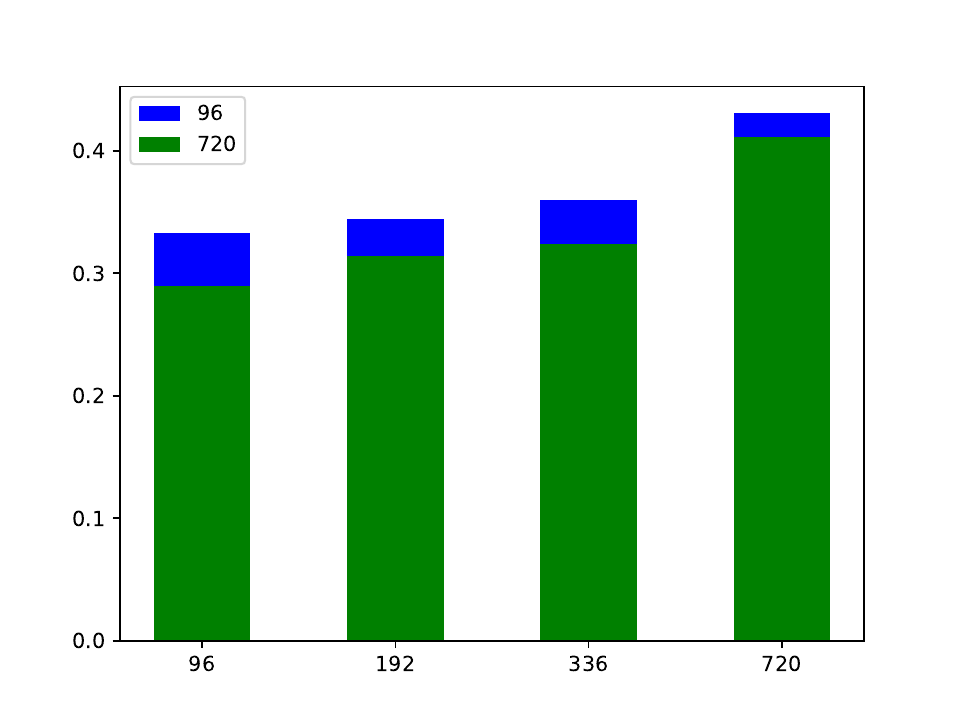}}
    \subfigure[\scalebox{0.8}{}] {\includegraphics[width=.25\textwidth]{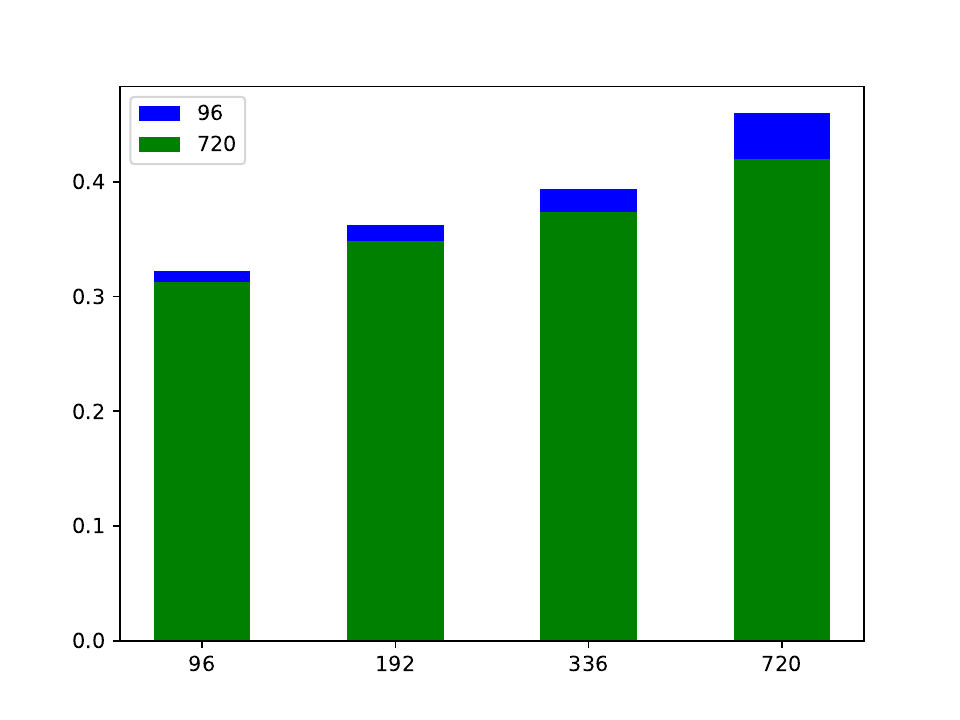}}
    \subfigure[\scalebox{0.8}{}] {\includegraphics[width=.25\textwidth]{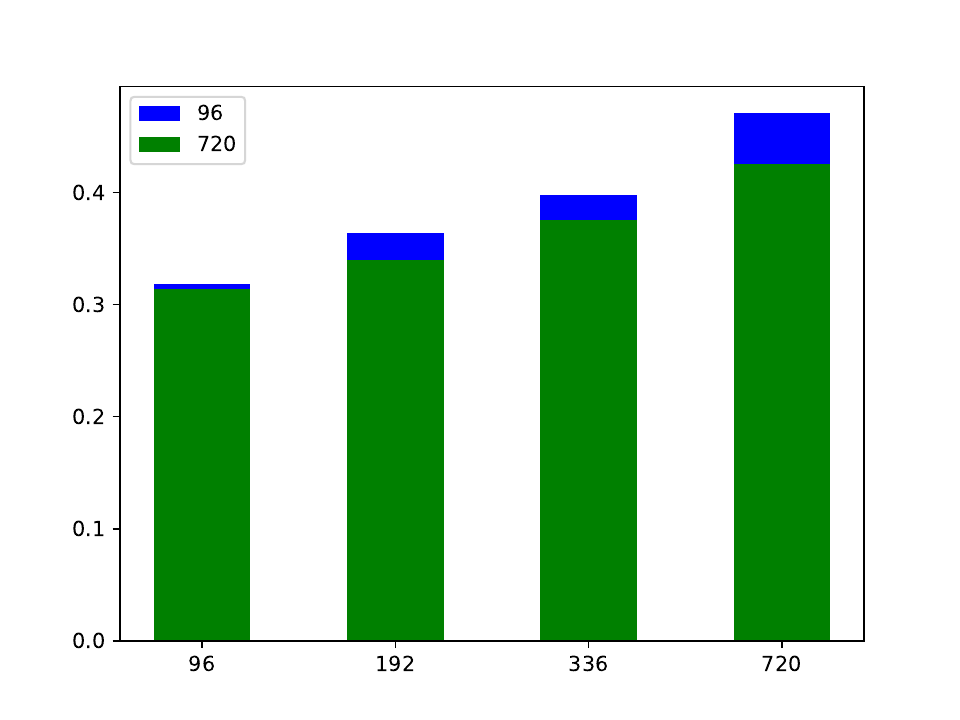}}
	\caption{The effect of look-back length: The 720 window size look-back (green) significantly reduces the prediction error compared to the 96 window size (blue). (a) The MSE of SCFormer-\textit{conv} on Solar-Energy. (b) The MSE of SCFormer-\textit{triangular} on Solar-Energy. (c) The MAE of SCFormer-\textit{conv} on Traffic. (d) The MAE of SCFormer-\textit{triangular} on Traffic. (e) The MSE of SCFormer-\textit{conv} on ETTm1. (f) The MSE of SCFormer-\textit{triangular} on ETTm1.}
	\label{fig: length}
    \vspace{-5mm}
\end{figure}

\renewcommand{\dblfloatpagefraction}{.8}
\begin{table}[htpb]
  \vskip 0.05in
  \centering
  \scalebox{0.55}{
  \begin{threeparttable}
  \begin{small}
  \renewcommand{\multirowsetup}{\centering}
  \setlength{\tabcolsep}{4.1pt}
  \begin{tabular}{c|c|c|cc|cc|cc|cc|cc}
    \toprule
    \multicolumn{3}{c|}{\multirow{2}{*}{{Models}}} & 
    \multicolumn{2}{c}{\rotatebox{0}{\scalebox{1.0}{Transformer}}} &
    \multicolumn{2}{c}{\rotatebox{0}{\scalebox{1.0}{Reformer}}} &
    \multicolumn{2}{c}{\rotatebox{0}{\scalebox{1.0}{Informer}}} &
    \multicolumn{2}{c}{\rotatebox{0}{\scalebox{1.0}{Flowformer}}} &
    \multicolumn{2}{c}{\rotatebox{0}{\scalebox{1.0}{Flashformer}}} \\
    \multicolumn{3}{c|}{}
    &\multicolumn{2}{c}{\scalebox{1.0}{}} & 
    \multicolumn{2}{c}{\scalebox{1.0}{}} & 
    \multicolumn{2}{c}{\scalebox{1.0}{}} & 
    \multicolumn{2}{c}{\scalebox{1.0}{}} & 
    \multicolumn{2}{c}{\scalebox{1.0}{}} \\
    \cmidrule(lr){4-5} \cmidrule(lr){6-7}\cmidrule(lr){8-9} \cmidrule(lr){10-11} \cmidrule(lr){12-13}  
    \multicolumn{3}{c|}{Metric}  & \scalebox{1.0}{MSE} & \scalebox{1.0}{MAE}  & \scalebox{1.0}{MSE} & \scalebox{1.0}{MAE}  & \scalebox{1.0}{MSE} & \scalebox{1.0}{MAE}  & \scalebox{1.0}{MSE} & \scalebox{1.0}{MAE} & \scalebox{1.0}{MSE} & \scalebox{1.0}{MAE}\\
    \toprule
    \multirow{10}{*}{\scalebox{1.0}{ECL}} & \multirow{5}{*}{Original} & 96 & 0.148 & 0.240 & 0.182 & 0.275 & 0.190 & 0.286 & 0.183 & 0.267 & 0.178 & 0.265 \\
    & & 192 & 0.162 & 0.253 & 0.192 & 0.286 & 0.201 & 0.297 & 0.192 & 0.277 & 0.189 & 0.276 \\
    & & 336 & 0.178 & 0.269 & 0.210 & 0.304 & 0.218 & 0.315 & 0.210 & 0.295 & 0.207 & 0.294 \\
    & & 720 & 0.225 & 0.317 & 0.249 & 0.339 & 0.255 & 0.347 & 0.255 & 0.332 & 0.251 & 0.329 \\
    \cmidrule(lr){3-13}
    & & Avg  & \scalebox{1.0}{0.178} & \scalebox{1.0}{0.270}& \scalebox{1.0}{0.208} & \scalebox{1.0}{0.301} & \scalebox{1.0}{0.216} & \scalebox{1.0}{0.311}  & \scalebox{1.0}{0.210} & \scalebox{1.0}{0.293} & \scalebox{1.0}{0.206} & \scalebox{1.0}{0.291}\\
    \cmidrule(lr){2-13}
    & \multirow{5}{*}{+HiPPO} & 96 & 0.129 & 0.228 & 0.144 & 0.241 & 0.144 & 0.241 & 0.142 & 0.239 & 0.140 & 0.239 \\
    & & 192 & 0.147 & 0.245 & 0.158 & 0.254 & 0.157 & 0.253 & 0.157 & 0.252 & 0.156 & 0.253 \\
    & & 336 & 0.160 & 0.260 & 0.173 & 0.270 & 0.171 & 0.268 & 0.172 & 0.269 & 0.171 & 0.271 \\
    & & 720 & 0.191 & 0.286 & 0.208 & 0.302 & 0.204 & 0.298 & 0.205 & 0.300 & 0.207 & 0.305 \\
    \cmidrule(lr){3-13}
    & & Avg  & \textbf{\scalebox{1.0}{0.156}} & \textbf{\scalebox{1.0}{0.254}} & \textbf{\scalebox{1.0}{0.170}} & \textbf{\scalebox{1.0}{0.266}} & \textbf{\scalebox{1.0}{0.169}} & \textbf{\scalebox{1.0}{0.265}} & \textbf{\scalebox{1.0}{0.169}} & \textbf{\scalebox{1.0}{0.265}} & \textbf{\scalebox{1.0}{0.168}} & \textbf{\scalebox{1.0}{0.267}} \\
    \midrule
    \multirow{10}{*}{\scalebox{1.0}{Traffic}} & \multirow{5}{*}{Original} & 96 & 0.395 & 0.268 & 0.617 & 0.356 & 0.632 & 0.367 & 0.493 & 0.339 & 0.464 & 0.320 \\
    & & 192 & 0.417 & 0.276 & 0.629 & 0.361 & 0.641 & 0.370 & 0.506 & 0.345 & 0.479 & 0.326 \\
    & & 336 & 0.433 & 0.283 & 0.648 & 0.370 & 0.663 & 0.379 & 0.526 & 0.355 & 0.501 & 0.337 \\
    & & 720 & 0.467 & 0.302 & 0.694 & 0.394 & 0.713 & 0.405 & 0.572 & 0.381 & 0.524 & 0.350 \\
    \cmidrule(lr){3-13}
    & & Avg  & \textbf{\scalebox{1.0}{0.428}} & \textbf{\scalebox{1.0}{0.282}} & \scalebox{1.0}{0.647} & \scalebox{1.0}{0.370} & \scalebox{1.0}{0.662} & \scalebox{1.0}{0.380} & \textbf{\scalebox{1.0}{0.524}} & \scalebox{1.0}{0.355} & \textbf{\scalebox{1.0}{0.492}} & \textbf{\scalebox{1.0}{0.333}} \\
    \cmidrule(lr){2-13}
    & \multirow{5}{*}{+HiPPO} & 96 & 0.448 & 0.333 & 0.558 & 0.357 & 0.586 & 0.359 & 0.566 & 0.332 & 0.531 & 0.350 \\
    & & 192 & 0.440 & 0.314 & 0.538 & 0.334 & 0.558 & 0.356 & 0.542 & 0.331 & 0.519 & 0.328 \\
    & & 336 & 0.521 & 0.360 & 0.543 & 0.342 & 0.575 & 0.347 & 0.549 & 0.342 & 0.531 & 0.340 \\
    & & 720 & 0.630 & 0.431 & 0.590 & 0.359 & 0.621 & 0.371 & 0.600 & 0.357 & 0.582 & 0.364 \\
    \cmidrule(lr){3-13}
    & & Avg  & \scalebox{1.0}{0.509} & \scalebox{1.0}{0.359} & \textbf{\scalebox{1.0}{0.557}} & \textbf{\scalebox{1.0}{0.348}} & \textbf{\scalebox{1.0}{0.585}} & \textbf{\scalebox{1.0}{0.358}} & \scalebox{1.0}{0.564} & \textbf{\scalebox{1.0}{0.340}} & \scalebox{1.0}{0.540} & \scalebox{1.0}{0.345} \\
    \midrule
    \multirow{10}{*}{\scalebox{1.0}{Weather}} & \multirow{5}{*}{Original} & 96 & 0.174 & 0.214 & 0.169 & 0.225 & 0.180 & 0.251 & 0.183 & 0.223 & 0.177 & 0.218 \\
    & & 192  & 0.221 & 0.254 & 0.213 & 0.265 & 0.244 & 0.318 & 0.231 & 0.262 & 0.229 & 0.261 \\
    & & 336 & 0.278 & 0.296 & 0.268 & 0.317 & 0.282 & 0.343 & 0.286 & 0.301 & 0.283 & 0.300 \\
    & & 720 & 0.358 & 0.349 & 0.340 & 0.361 & 0.377 & 0.409 & 0.363 & 0.352 & 0.359 & 0.251 \\
    \cmidrule(lr){3-13}
    & & Avg  &  \scalebox{1.0}{0.258} & \scalebox{1.0}{0.279} & \scalebox{1.0}{0.248} & \scalebox{1.0}{0.292} & \scalebox{1.0}{0.271} & \scalebox{1.0}{0.330} & \scalebox{1.0}{0.266} & \scalebox{1.0}{0.285} & \scalebox{1.0}{0.262} & \scalebox{1.0}{0.282} \\
    \cmidrule(lr){2-13}
    & \multirow{5}{*}{+HiPPO} & 96 & 0.156 & 0.205 & 0.164 & 0.212 & 0.162 & 0.211 & 0.165 & 0.212 & 0.168 & 0.215 \\
    & & 192 & 0.212 & 0.254 & 0.210 & 0.253 & 0.211 & 0.254 & 0.209 & 0.252 & 0.207 & 0.253 \\
    & & 336 & 0.261 & 0.293 & 0.260 & 0.293 & 0.260 & 0.292 & 0.257 & 0.289 & 0.263 & 0.293 \\
    & & 720 & 0.313 & 0.334 & 0.328 & 0.337 & 0.315 & 0.337 & 0.326 & 0.337 & 0.320 & 0.338 \\
    \cmidrule(lr){3-13}
    & & Avg  &\textbf{\scalebox{1.0}{0.235}} & \textbf{\scalebox{1.0}{0.271}} & \textbf{\scalebox{1.0}{0.240}} & \textbf{\scalebox{1.0}{0.273}} & \textbf{\scalebox{1.0}{0.237}} & \textbf{\scalebox{1.0}{0.273}} & \textbf{\scalebox{1.0}{0.239}} & \textbf{\scalebox{1.0}{0.272}} & \textbf{\scalebox{1.0}{0.239}} & \textbf{\scalebox{1.0}{0.274}} \\
    \bottomrule
  \end{tabular}
    \end{small}
  \end{threeparttable}}
  \caption{The results of variant Transformers equipped with cumulative historical state.}
  \label{tab:full_forecasting_promotion}
  \vspace{-10mm}
\end{table}

\begin{figure}[htpb]
\vspace{-5mm}
\centering 
\subfigure[]{
\label{fig: heatmap}
\includegraphics[scale=0.12]{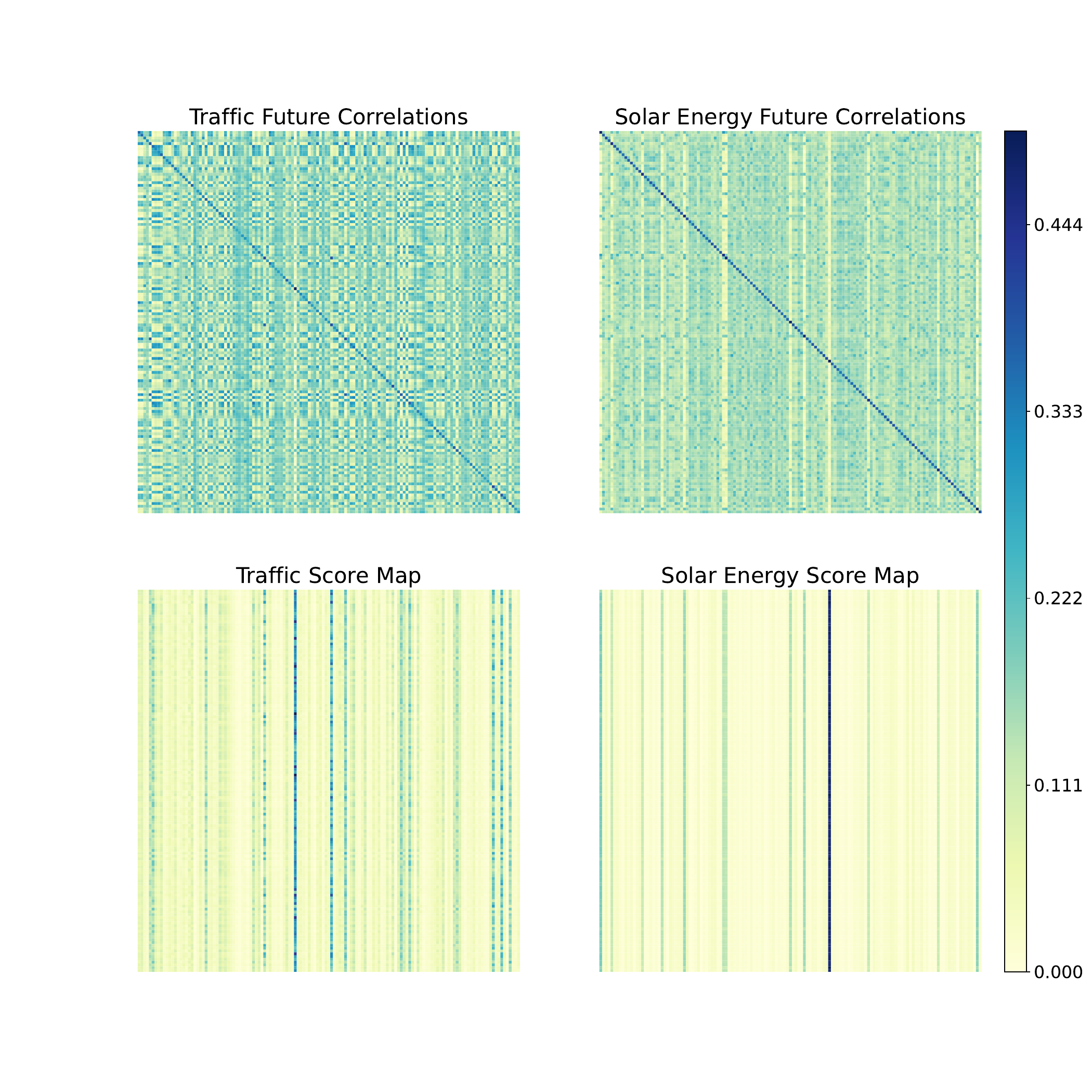}}
\subfigure[]{
\label{Fig.ecl.itrans}
\includegraphics[scale=0.25]{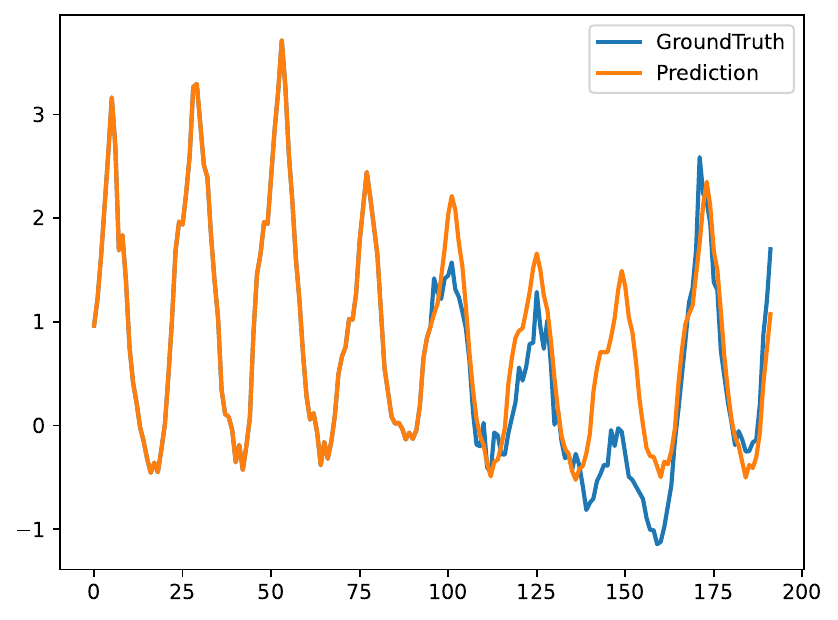}}
\subfigure[]{
\label{Fig.ecl.ours}
\includegraphics[scale=0.25]{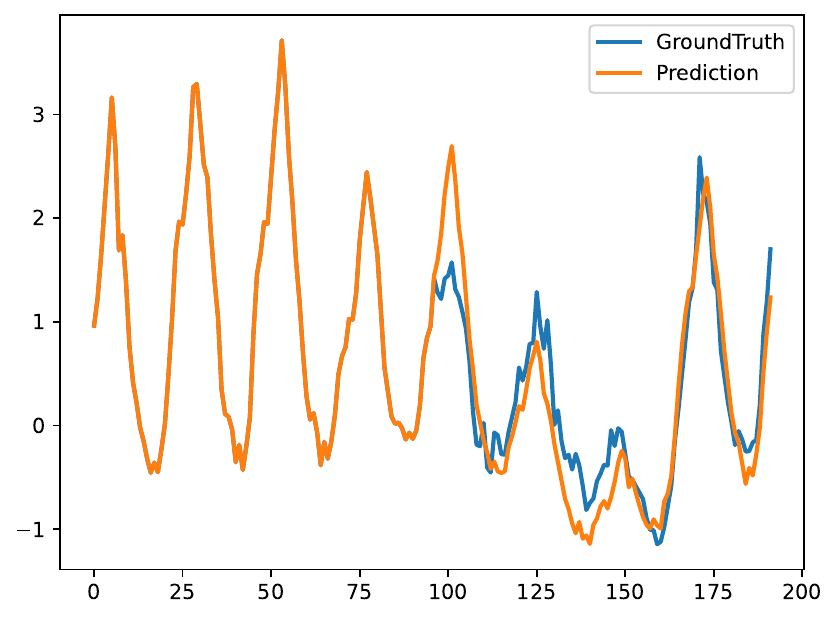}}
\caption{(a) Channels(multivariate) correlations: Left-Top: the future correlations of \textit{Traffic}; Left-Bottom: the attention scores of \textit{Traffic}; Right-Top: the future correlations of \textit{Solar-Energy}; Right-Bottom: the attention scores of \textit{Solar-Energy}. (b) Example visualization of iTransformer on ECL. (c) Example visualization of SCFormer-\textit{triangular} on ECL.}
\label{fig: visual_ecl}
\vspace{-5mm}
\end{figure}

\subsection{Model Analysis}
\subsubsection{HiPPO with Variant Transformers}
We apply HiPPO to other variants of Transformers, including Reformer~\cite{kitaev2020reformer}, Informer~\cite{zhou2021informer}, Flowformer~\cite{wu2022flowformer}, and Flashformer~\cite{dao2022flashattention}, with the results shown in Table \ref{tab:full_forecasting_promotion}. By incorporating the channel-wise strategy, HiPPO improves performance in most cases, demonstrating its effectiveness in maintaining the state of long historical information across different models.
\vspace{-4mm}

\subsubsection{Length Effect of Look-back}
\label{c1}
We examine whether increasing the look-back length further improves performance. As shown in Figure \ref{fig: length}, the model's performance continues to improve with a longer look-back, indicating that HiPPO-based cumulative historical state and the look-back are decoupled. The look-back captures short-term changes, while the cumulative historical state encodes global features from the more entire historical series.
\vspace{-4mm}

\subsubsection{Case Study}
To intuitively illustrate the advantages of our method, we compare it with iTransformer using an example from the \textit{ECL} dataset. Figure \ref{fig: visual_ecl}(b-c) shows that iTransformer's prediction curve is significantly distorted around timestamp 150, while our method provides more accurate predictions. We also plot the attention scores in the model's last layer and the future correlations for the \textit{Traffic} and \textit{Solar-Energy} datasets in Figure \ref{fig: heatmap}. The results show that the model is able to clearly learn the correlations between channels within the prediction horizon, for example, brighter columns in future correlations correspond to darker areas in the attention scores. However, compared to the \textit{Solar} dataset, the patterns in the \textit{Traffic} dataset are less pronounced, which indirectly explains why the model performs less optimally on the \textit{Traffic} dataset.

\section{Conclusion}
In this paper, we propose SCFormer, a multivariate time series forecasting model. SCFormer uses 1D convolutions and triangular matrices to structure the linear transformations in the channel-wise Transformer, thereby introducing temporal constraints. Additionally, we introduce a method for maintaining the cumulative historical state based on HiPPO, which serves as a simple and efficient memory mechanism, allowing the model to capture historical information beyond the fixed look-back window. Extensive comparative experiments, ablation studies, and analytical evaluations confirm the effectiveness of the proposed method.

%
%
%
\bibliographystyle{splncs04}
\bibliography{sample-base}
%




\end{document}